\pdfoutput=1
\documentclass{article}

\usepackage[preprint]{neurips_2026}

\usepackage[utf8]{inputenc}
\usepackage[T1]{fontenc}
\usepackage{hyperref}
\usepackage{url}
\usepackage{booktabs}
\usepackage{amsfonts}
\usepackage{nicefrac}
\usepackage{microtype}
\usepackage{xcolor}
\usepackage{inconsolata}

\usepackage{graphicx}
\usepackage{multirow}
\usepackage{tabularx}
\usepackage{amsmath}

\title{From Skill Text to Skill Structure: \\
The Scheduling-Structural-Logical Representation\\
for Agent Skills}

\author{
  Qiliang Liang$^{1,2}$, Hansi Wang$^{1,2}$, Zhong Liang$^3$, and Yang Liu$^{1,2}$\thanks{Corresponding author.} \\
  $^1$Key Laboratory of Computational Linguistics, Ministry of Education, Peking University\\
  $^2$School of Computer Science, Peking University\\
  $^3$Department of Chinese Language and Literature, Peking University\\
  \texttt{\{lql.pkucs,lzliangzh\}@gmail.com, \{wanghansi2019,liuyang\}@pku.edu.cn}\\
}
\begin{document}
\maketitle
\setcounter{footnote}{0}

\begin{abstract}
Large language model (LLM) agents increasingly rely on reusable \textit{skills}: capability packages that combine instructions, control flow, constraints, and tool calls. 
In current agent systems, however, skills are still represented by text-heavy artifacts, mainly \texttt{SKILL.md}-style documents whose machine-usable evidence remains embedded largely in natural-language descriptions.
As a result, skill-centered agent systems face a representation problem: both managing skill collections and using skills during agent execution require reasoning over invocation interfaces, execution structure, and concrete side effects, but these signals are often entangled in a single textual surface. 
An explicit representation of skill knowledge may therefore help make these artifacts easier for machines to acquire and leverage. Drawing on Memory Organization Packets, Script Theory, and Conceptual Dependency from Schank and Abelson's classical work on cognitive linguistic representation, we introduce what is, to our knowledge, the first structured representation for agent skill artifacts that disentangles skill-level scheduling signals, scene-level execution structure, and logic-level action/resource-use evidence: the Scheduling-Structural-Logical (SSL) representation.
We instantiate SSL with an LLM-based normalizer and evaluate SSL-derived representations in two tasks, Skill Discovery and Risk Assessment.
The experiment shows that SSL significantly outperforms the text-only baselines: in Skill Discovery, MRR@50 improves from 0.649 to 0.729; in Risk Assessment, macro F1 improves from 0.409 to 0.509. These findings suggest that an explicit, source-grounded structure can make agent skills easier to search and review, positioning SSL as a practical step toward more inspectable, reusable, and operationally actionable skill representations, rather than a finished standard or end-to-end skill-management mechanism.
\footnote{The SSL guidelines, annotated skill corpus, and evaluation datasets are available at \url{https://github.com/COOLPKU/SSL}.}
\end{abstract}

\section{Introduction}
\label{sec:introduction}

Large language models (LLMs) increasingly serve as controllers for agent systems that maintain task context, coordinate multi-step workflows, and act through external resources such as files, tools, or services \citep{xi2023rise_llm_agents, luo2025llm_agent_survey}. As these agent systems move beyond isolated tool calls, reusable capabilities are increasingly packaged as \textit{skills}: bundles of instructions, control flow, constraints, and callable operations that can be discovered, selected, governed, and reused across tasks \citep{evolution_tool_use_2026, sok_agentic_skills_2026, skillnet_2026}.

Despite this shift, skills are still usually exposed through text-dominant artifacts: \texttt{SKILL.md}-style instruction files and README-like documentation, sometimes wrapped in JSON/YAML records whose machine-usable evidence remains in natural-language fields. This reflects a long-standing trade-off in machine-consumable documentation: natural language is easy for people to author and read, but difficult for automated systems to analyze, validate, and reuse reliably \citep{bernerslee2001semanticweb, gonzalezmora2023openapi_documentation, lazar2025oasbuilder}. A single skill artifact can therefore entangle a skill's invocation interface, execution phases, and action/resource-use evidence, forcing downstream components to infer these properties from long, noisy, or potentially incomplete text.

This results in a fundamental representational bottleneck of skills across downstream uses: semantically distinct properties are collapsed into a single textual surface. For example, for discovery of relevant skills, large and overlapping registries require signals beyond sparse metadata, including implementation-level cues needed for selection \citep{skillrouter_2026, graph_of_skills_2026}; for assessment of pre-execution risk, third-party skills may be installed with broad or persistent access, yet their instructions, configuration files, and executable operations are often inspected together, obscuring risks such as data exfiltration and privilege escalation \citep{agent_skills_wild_2026, secure_agent_skills_2026, skillsieve_2026}. These observations point to a missing layer between raw skill artifacts and downstream systems: a reusable, source-grounded representation that can expose skill evidence across tasks without repeatedly re-parsing the full instruction text.

\begin{figure}[t]
	\centering
	\includegraphics[width=\linewidth]{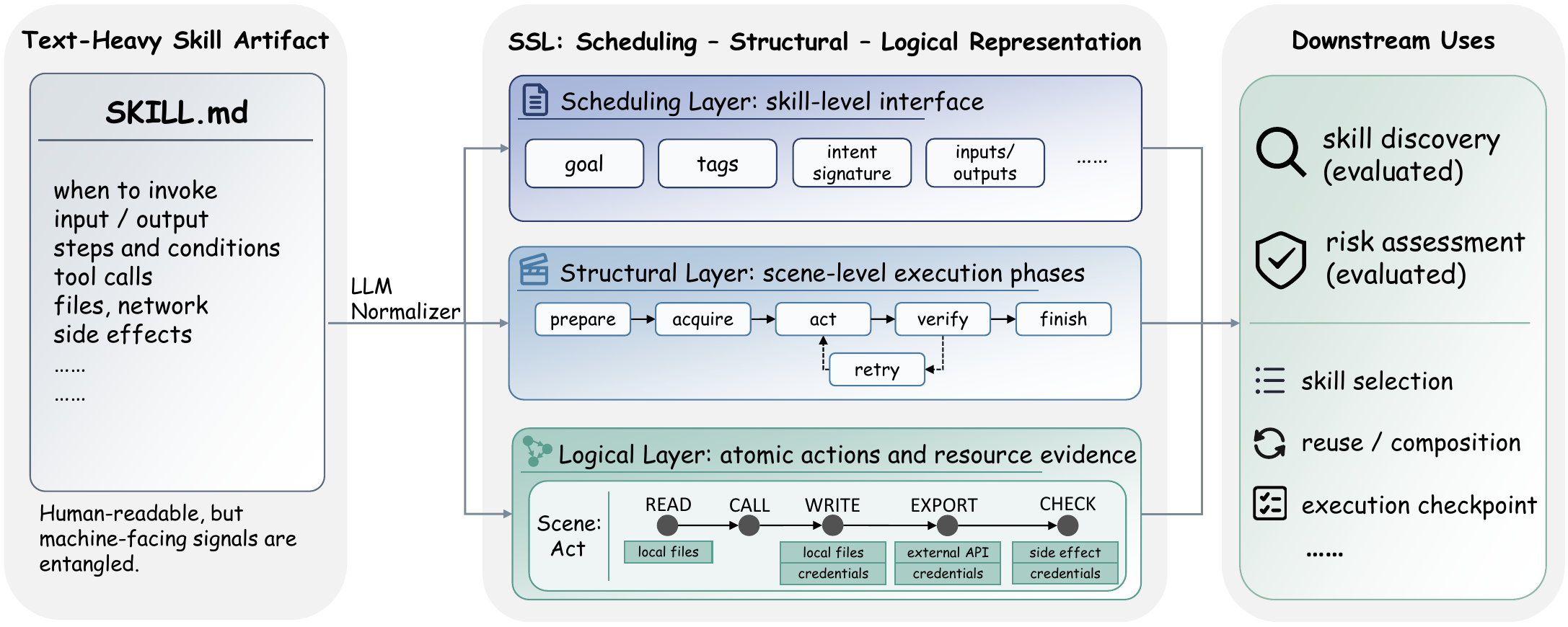}
	\caption{Overview of the SSL representation. A text-heavy skill artifact is converted by a normalizer into three layers: a scheduling record for invocation-level signals, a structural graph of execution scenes, and a logical graph of atomic actions and resource-use evidence. The structured view remains paired with the original source document and supports downstream tasks such as Skill Discovery and Risk Assessment.}
	\label{fig:ssl-overview}
\end{figure}

To address this gap, we propose the \textbf{Scheduling-Structural-Logical (SSL)} representation. To our knowledge, it is the first structured representation designed specifically for disentangling agent skill artifacts. SSL maps an unstructured skill document into a typed three-layer JSON graph organized around three analogies from Schank and Abelson's classical work on cognitive linguistic representation: The \textbf{Scheduling} layer draws on Memory Organization Packets as goal-oriented organizers for retrieving and contextualizing experience \citep{schank1980language_memory}; The \textbf{Structural} layer draws on Script Theory, which represents stereotyped activities as ordered scenes with expectations and transitions \citep{schank1977scripts}; The \textbf{Logical} layer draws on Conceptual Dependency, which decomposes linguistic meaning into primitive action structures that abstract away from surface wording \citep{schank1972conceptual_dependency}. Together, these theories provide a reference point for disentangling skills into goal-level context, ordered execution trajectory, and primitive operations. Inspired by this reference point, SSL is hence designed to represent machine-facing skill artifacts.
An overview of the resulting representation and its role in downstream skill-centered tasks is illustrated in Figure~\ref{fig:ssl-overview}.

We then instantiate SSL with an LLM-based normalizer that converts existing \texttt{SKILL.md} files into the SSL schema. While SSL is intended as a general intermediate representation for skill-centered agent systems, we evaluate it through two downstream settings that are both practically important and measurable offline: \textbf{Skill Discovery} tests whether interface-level and structural signals help match user requests to the right skill in a large registry; \textbf{Risk Assessment} tests whether action- and resource-level signals help reviewers identify operational risks. Across the evaluations, SSL significantly outperforms text-only skill representations: in {Skill Discovery}, a rich SSL-derived description view improves retrieval MRR@50 from 0.649 to 0.729; in {Risk Assessment}, the combined \texttt{SKILL.md} + SSL view improves macro F1 from 0.409 to 0.509 over the complete source document alone. These results reveal that SSL exposes useful evidence across distinct skill-centered tasks, while remaining complementary to the original source document.

To sum up, we make three contributions in this paper:
\begin{itemize}
	\item We introduce SSL, a three-layer representation for agent skill artifacts that disentangles skill-level scheduling signals, scene-level execution structure, and logic-level action/resource-use evidence, and instantiate it for existing \texttt{SKILL.md} files with an LLM-based normalizer;
	\item We evaluate SSL in two skill-centered tasks and gain significant outperformance over text-only baselines: in Skill Discovery, MRR@50 improves from 0.649 to 0.729; in Risk Assessment, macro F1 improves from 0.409 to 0.509;
	\item We present release-ready evaluation datasets over public agent skills, including a 6,184-skill corpus, 431 intent-level queries for Skill Discovery, and 252 skills with six-dimensional Risk Assessment labels.
\end{itemize}

\section{Related Work}

\subsection{LLM Agents and the Rise of Reusable Skills}
Nowadays, research on LLMs has increasingly moved beyond standalone prediction and generation toward agent systems that maintain task context, plan over multiple steps, and act through external tools \citep{xi2023rise_llm_agents, luo2025llm_agent_survey}. A first line of this work studied tool use, where external capabilities were often modeled as atomic APIs or functions to be selected, called, and incorporated into model reasoning \citep{toolformer, gorilla, toolllm}. That framing is natural for single-tool invocation, but less adequate when an agent capability includes instructions, control flow, constraints, resources, and executable operations. Systems such as Voyager \citep{voyager_2023} point toward this higher-level abstraction by maintaining a library of executable skills that can be retrieved and reused across tasks. Recent work has extended this view to skill repositories, trajectory-derived skill construction, inference-time routing, and training-time internalization \citep{sok_agentic_skills_2026, skillnet_2026, skillrouter_2026, skillx_kb_2026, skill0_2026}.

These efforts mostly treat the representation of an individual skill as an implicit substrate: a repository entry, a learned unit, or a routing target. The question left open is how an existing skill should be exposed in a machine-usable form that disentangles invocation interface, scene-level execution structure, and action/resource-use evidence.

\subsection{Linguistic Knowledge Representations of Activities}
Linguistic knowledge representation studies how recurring activities can be organized beyond surface text. Within Schank and Abelson's line of classical work, Memory Organization Packets model recurring goal-oriented contexts \citep{schank1980language_memory}; Script Theory represents stereotyped activities as ordered event sequences with roles and transitions \citep{schank1977scripts}; and Conceptual Dependency decomposes linguistic meaning into primitive action structures \citep{schank1972conceptual_dependency}. Related frame-based theories make a similar commitment to structured context: Minsky's Frame Theory represents familiar situations through slots, defaults, and expectations \citep{minsky1975framework}, and Fillmore's Frame Semantics treats meaning as grounded in scenes with participant roles \citep{fillmore1982frame_semantics}.

While frame-based theories may support the broader premise that linguistic content can be organized around context, Schank and Abelson's work provides a closer analogy for SSL by distinguishing goal-oriented contexts, ordered activities, and primitive operations, aligning closely with the three kinds of evidence SSL disentangles: invocation-level interfaces, scene-level execution structure, and atomic action/resource-use evidence.

\subsection{Skill Retrieval and Routing}
Selecting a skill from a collection requires matching a user request to one or more executable capabilities. This setting resembles query--document retrieval, but each candidate is an executable capability rather than an ordinary passage. Neural retrieval work has shown that representation quality is central to query--candidate matching \citep{reimers2019sentencebert, karpukhin2020dpr, thakur2021beir}. Tool- and skill-retrieval work further shows that capability matching depends on signals scattered across interface fields, instructional text, examples, implementation details, and structural dependencies, and that generic retrievers do not always transfer reliably to tool or skill selection \citep{craft_2024, toolret_2025, toolrerank_2024, masstool_2025, skillrouter_2026, graph_of_skills_2026}. Related work on tool-document compression and enrichment similarly argues that verbose tool descriptions often need to be reorganized for effective retrieval \citep{easytool_2025, toolrex_2026}.

These studies motivate better retrieval models and more effective tool descriptions, but they still largely assume that candidate representations are already available. A remaining question is how each skill should be represented before retrieval, so that matching may leverage explicit interface, structural, and operational cues rather than relying on raw documentation alone.

\subsection{Security and Risk for Tool-Using Agents}

Security risks in tool-using agents often arise at the boundary between natural-language instructions and external capabilities. Indirect prompt-injection work shows that retrieved or tool-returned text can blur the distinction between data and instructions, while agent-security benchmarks extend this concern to multi-step settings with tools, memory, untrusted observations, and harmful user goals \citep{greshake2023indirect_prompt_injection, toolemu_2024, agentdojo_2024, asb_2025, agentharm_2025}. Work on prompt-flow integrity, mandatory access control, and least-authority execution frames safety as a question of privilege boundaries and resource access \citep{pfi_2025, seagent_2026}. Skill-specific security studies further show that reusable skills can become an attack surface or review target because natural-language instructions, executable code, implicit trust, and distribution-time reuse make side effects difficult to inspect from text alone \citep{skillattack_2026, agent_skills_wild_2026, secure_agent_skills_2026, skillsieve_2026}.

For security analysis, the remaining challenge is to make risk-relevant evidence inspectable before a skill is invoked. Such evidence includes action types, resource scopes, dependencies, and data-flow cues that are often distributed across natural-language instructions.

\section{The Scheduling-Structural-Logical Representation of Agent Skills}
\label{sec:methodology}

SSL represents a skill artifact by disentangling three kinds of source-grounded evidence: when the skill should be invoked, how its work is organized, and what operations or resources it may involve. This decomposition is inspired by Schank and Abelson's cognitive linguistic representation theories: Memory Organization Packets motivate goal- and context-level capability records \citep{schank1980language_memory}; Script Theory motivates ordered execution phases with conditions and transitions \citep{schank1977scripts}; and Conceptual Dependency motivates primitive action structures with roles, effects, and resource targets \citep{schank1972conceptual_dependency}. These theories serve as design analogies for a systems schema whose purpose is to make skill artifacts easier to manage, inspect, validate, and reuse.

\subsection{Problem Formulation and Design Goals}
Let $d$ denote a skill artifact, such as a \texttt{SKILL.md} file. SSL maps $d$ into a typed representation:
\begin{equation}
G_d = (r_{\mathrm{sch}}, G_{\mathrm{str}}, G_{\mathrm{log}}, R_{\mathrm{cont}}, R_{\mathrm{entry}}),
\label{eq:ssl-representation}
\end{equation}
where $r_{\mathrm{sch}}$ is the \textit{Scheduling} layer, a skill-level interface record that captures invocation signals; $G_{\mathrm{str}}$ is the \textit{Structural} layer, a scene-level graph that captures execution phases and their transitions; $G_{\mathrm{log}}$ is the \textit{Logical} layer, a logic-step graph that captures atomic actions and resource-use evidence; $R_{\mathrm{cont}}$ records containment across layers; and $R_{\mathrm{entry}}$ records entry pointers.

Operationally, $r_{\mathrm{sch}}$ is realized as a single skill-level record, while the remaining two layers are realized as record graphs: Scene records instantiate $G_{\mathrm{str}}$ as a directed graph over execution phases; and logic-step records instantiate $G_{\mathrm{log}}$ as a directed graph over atomic actions within those phases. The two auxiliary relations specify how these records are assembled: $R_{\mathrm{cont}}$ assigns scenes to the skill and logic steps to scenes; while $R_{\mathrm{entry}}$ identifies where traversal begins. The complete field-level realization is listed in Appendix~\ref{app:ssl-schema}.

SSL follows three design goals: it is \textit{compact}, preserving evidence needed for skill management and use while avoiding open-ended or subjective attributes; it is \textit{typed}, using restricted vocabularies so normalized outputs remain comparable across skills; and it is \textit{grounded}, that fields strictly summarize evidence present in the source artifact, making no attempt to infer hidden behavior. Appendix~\ref{app:ssl-example} provides a compact complete instance.

\subsection{The Three-Layer SSL Representation}

As illustrated in Figure~\ref{fig:ssl-overview}, we detail the three-layer SSL representation as follows.

\subsubsection{Scheduling Layer: Skill-Level Interface}
The scheduling layer corresponds to $r_{\mathrm{sch}}$ in Eq.~\ref{eq:ssl-representation}. Instead of treating a skill merely as an instruction document, this layer approaches the artifact as an invocation-level capability unit. It exposes supported intents, input/output contracts, and coarse dependencies or control-flow properties before deeper inspection. This gives each skill a stable capability record that can be compared across a repository without unfolding its full scene or logic-step structure.

\subsubsection{Structural Layer: Scene-Level Execution Phases}
The structural layer corresponds to $G_{\mathrm{str}}$ in Eq.~\ref{eq:ssl-representation}. Its nodes are scenes, and its edges represent phase-level transitions among those scenes. Motivated by script-like event structure \citep{schank1977scripts}, this layer groups low-level operations into coherent stages, such as preparation, acquisition, reasoning, action, verification, and recovery, making the skill's phase organization visible before the reader inspects individual logic steps.

\subsubsection{Logical Layer: Atomic Actions and Resource Evidence}
The logical layer corresponds to $G_{\mathrm{log}}$ in Eq.~\ref{eq:ssl-representation}. Its nodes are logic steps, and its edges represent micro-level transitions among source-grounded atomic actions. Each atomic action selects an \texttt{act\_type} from the closed primitive inventory as described in Appendix~\ref{app:ssl-schema}, and records arguments, effects, and resource boundaries as typed evidence. Because atomicity is a property of the representation irrespective of runtime details, a logic step is simply the smallest operational unit the source artifact supports without inventing missing implementation details.

\subsection{The SSL Normalization Pipeline}
We instantiate SSL with an LLM-based normalizer implemented with DeepSeek-V3.2~\citep{liu2025deepseek}. It converts the full source document of the skill artifact into the SSL representation. Operating strictly as a semantic extractor, the normalizer avoids open-ended summarization: its prompt specifies the schema, allowed vocabularies, and grounding policy, and every populated field must be supported by the source artifact. As summarized in Appendix~\ref{app:normalizer-prompt}, the pipeline extracts the skill-level record, decomposes the document into scenes, expands each scene into source-grounded logic steps, and validates the resulting graph. Validation checks structural well-formedness, identifier consistency, allowed enum values, containment links, entry pointers, and transition targets. Outputs that fail parsing or hard validation are retried. To check whether this LLM-based normalizer produces source-grounded outputs, we conduct a human audit of 100 annotated skills, with 83\% of audited SSL judged to be supported by the corresponding source artifacts (Appendix~\ref{app:normalizer-human-audit}).

\section{Evaluation}
\label{sec:evaluation}
We evaluate SSL as an intermediate representation for skill-centered agent systems through two downstream tasks: Skill Discovery and Risk Assessment. The first asks whether a compact structured view helps route user requests to the correct skill in a large registry; the second asks whether it helps an LLM judge recover risk signals that are easier to miss in text-only representations. These tasks are not meant to exhaust the uses of SSL: they are selected to test whether the representation exposes useful interface-level and operation-level evidence under controlled comparisons.

\subsection{Evaluation I: Skill Discovery}
\subsubsection{Benchmark Construction}
We collect and formalize 6,184 public skills as the retrieval candidate pool. The main test set contains 431 intent-level requests. Each query is paired with its source skill, which is treated as the single relevant answer. The retained queries cover functional, constraint-based, compositional, safety-oriented, and scenario-style requests, while excluding weakly grounded queries, source-document paraphrases, and queries that reveal the skill name. Appendix~\ref{app:retrieval-by-type} details the generation, annotation and filtering.

\subsubsection{Input Representations and Evaluation Protocol}
We compare ten retrieval inputs under the same embedding and ranking pipeline. They consist of two text-only baselines, two source-outline controls, and six SSL-augmented variants. The text-only baselines use either the short metadata description or the complete \texttt{SKILL.md}. The source-outline controls add a generic compressed outline of the source document, testing whether SSL gains are reducible to source-document compression. The SSL variants pair the same two source contexts with three levels of structured augmentation. The ten settings are:
\begin{itemize}
	\item \textbf{Non-SSL baselines}: \textbf{Desc\_only} embeds the short natural-language description, \textbf{Full \texttt{SKILL.md}} embeds the complete source document, \textbf{Source Outline} embeds deterministic excerpts from the original \texttt{SKILL.md}, and \textbf{Desc + Source Outline} combines the description with those source-only excerpts;
	\item \textbf{SSL-Shallow variants}: \textbf{Desc + SSL-Shallow} and \textbf{Full \texttt{SKILL.md} + SSL-Shallow} add shallow normalized SSL fields: skill name, tags, and goal;
	\item \textbf{SSL-Sched variants}: \textbf{Desc + SSL-Sched} and \textbf{Full \texttt{SKILL.md} + SSL-Sched} add a compact scheduling view: skill name, goal, tags, intent signature, control-flow features, and an aggregate scene profile;
	\item \textbf{SSL-Rich variants}: \textbf{Desc + SSL-Rich} and \textbf{Full \texttt{SKILL.md} + SSL-Rich} add richer SSL-derived fields, including the skill identifier, explicit scene types and goals, dependencies, top pattern, and expected inputs and outputs.
\end{itemize}
All methods rank the same 6,184 candidates using a FAISS inner-product index over L2-normalized embeddings, and all dense vectors are produced with Qwen3-Embedding-0.6B~\citep{qwen3embedding_2025}. We report mean reciprocal rank within the top-50 retrieved candidates (MRR@50) as the primary metric because each query has one source skill, and use NDCG@5, NDCG@10, and Recall@10 to measure top-rank quality and coverage.

\subsubsection{Results and Analysis}

\begin{table}[h]
	\centering
	\small
		\begin{tabular}{p{0.17\textwidth}lcccc}
			\toprule
			\textbf{Group} & \textbf{Method} & \textbf{MRR@50} & \textbf{NDCG@5} & \textbf{NDCG@10} & \textbf{Recall@10} \\
			\midrule
			\multirow{4}{0.17\textwidth}{\textit{Baselines}} & Desc\_only & 0.588 & 0.608 & 0.626 & 0.761 \\
			 & Full \texttt{SKILL.md} & 0.645 & 0.655 & 0.682 & 0.821 \\
			 & Source Outline & 0.592 & 0.615 & 0.634 & 0.787 \\
			 & Desc + Source Outline & 0.649 & 0.670 & 0.689 & 0.833 \\
			\midrule
			\multirow{6}{0.17\textwidth}{\textit{SSL-augmented variants}} & Desc + SSL-Shallow & 0.716 & 0.730 & 0.752 & 0.879 \\
			 & Desc + SSL-Sched & 0.694 & 0.716 & 0.733 & 0.868 \\
			 & Desc + SSL-Rich & \textbf{0.729} & \textbf{0.748} & \textbf{0.770} & \textbf{0.905} \\
			 & Full \texttt{SKILL.md} + SSL-Shallow & 0.664 & 0.688 & 0.706 & 0.847 \\
			 & Full \texttt{SKILL.md} + SSL-Sched & 0.676 & 0.694 & 0.715 & 0.849 \\
			 & Full \texttt{SKILL.md} + SSL-Rich & 0.681 & 0.705 & 0.720 & 0.856 \\
			\bottomrule
		\end{tabular}
		\caption{Skill Discovery performance for 431 intent-level queries over the 6,184-skill corpus. All variants use the same embedding model and FAISS ranking pipeline; only the embedded skill representation changes.}
	\label{tab:retrieval}
\end{table}

The main result in Table~\ref{tab:retrieval} is that the best SSL-augmented retrieval input is \texttt{Desc + SSL-Rich}. It achieves the highest performance across all reported metrics, improving MRR@50 from 0.649 to 0.729 over \texttt{Desc + Source Outline}, the strongest non-SSL baseline. It also outperforms the complete source document, \texttt{Full \texttt{SKILL.md}} (0.645), indicating that the gain is not explained by simply embedding more source text.

The ablation further shows that the choice of structured fields matters: Shallow normalized fields already provide a strong gain over the raw description; while the compact scheduling view does not dominate the shallow variant; the richest SSL view performs best because it adds scene-level and interface-level signals, whereas full-document inputs remain weaker even when augmented with SSL. This suggests that concise structured representations are effective retrieval interfaces for intent-level skill search. As reported in Appendix~\ref{app:retrieval-by-type}, the same pattern remains visible when the results are broken down by query type. 
Appendix~\ref{app:bootstrap-ci} reports a paired bootstrap confidence interval for the main comparison: the MRR@50 improvement of \texttt{Desc + SSL-Rich} over \texttt{Desc + Source Outline} is 0.080, with a 95\% interval of [0.051, 0.111]

\subsection{Evaluation II: Risk Assessment}
\subsubsection{Benchmark Construction}
The Risk Assessment benchmark contains 252 gold-labeled skills sampled from the 6,184-skill corpus. Because many public skills expose little observable operational footprint, we use stratified sampling to improve coverage of risk-relevant evidence such as resource scopes, dependencies, tool calls, and side effects. Appendix~\ref{app:risk-assessment-benchmark} details the sampling procedure.

Each skill is labeled on six artifact-level dimensions: \textit{data exfiltration}, \textit{destructive behavior}, \textit{privilege escalation}, \textit{covert execution}, \textit{resource abuse}, and \textit{credential access}. These dimensions adapt recurring concerns from tool-use and agent-security work to evidence observable in skill artifacts, including resource scopes, dependencies, control-flow features, tool calls, and data-flow descriptions \citep{toolemu_2024, agentdojo_2024, asb_2025, agentharm_2025, agent_skills_wild_2026, pfi_2025, seagent_2026}. Each dimension receives a binary label, \textit{risk} or \textit{no risk}, based on observable static evidence. As listed in Appendix~\ref{app:risk-dimensions}, each dimension is accompanied by explicit boundary rules.

Gold labels are produced by a three-model pipeline using Gemini-3.1-pro-preview, Claude-Sonnet-4.5, and GPT-5 \citep{google_gemini3_docs_2026, anthropic_sonnet45_2025, openai_gpt5_system_card_2025}. Each model receives both the complete \texttt{SKILL.md} and the complete SSL representation, and assigns a binary label for each risk dimension. The SSL record is used as labeling evidence, but the label definition remains source-grounded: a positive label must be supported by observable evidence in the original skill artifact. Final labels are obtained through majority aggregation with disagreement review and sampled manual audit, as detailed in Appendix~\ref{app:risk-assessment-benchmark}.

\subsubsection{Input Representations and Evaluation Protocol}
For evaluation, we fix the judge to DeepSeek-V3.2 \citep{liu2025deepseek} and vary only the representation supplied to the judge. This isolates whether SSL changes the evidence available to the same evaluator. We compare five input representations including two baselines and three SSL-related variants, and report per-dimension F1 as well as macro-averaged accuracy, precision, recall, and F1 across the six dimensions:
\begin{itemize}
	\item \textbf{Desc Only}: the original registry name and description, without SSL-derived fields;
	\item \textbf{Full \texttt{SKILL.md}}: the complete source document, without SSL-derived fields;
	\item \textbf{SSL-Shallow}: normalized SSL interface fields, namely skill name, goal, and tags;
	\item \textbf{Full SSL}: the complete structured representation;
	\item \textbf{Full \texttt{SKILL.md} + SSL}: both the source document and the complete structured representation.
\end{itemize}

\subsubsection{Results and Analysis}

\begin{table}[h]
	\centering
	\small
		\begin{tabular}{lccccc}
			\toprule
			\textbf{Threat Dimension} & \textbf{Desc} & \textbf{Full MD} & \textbf{SSL-Sh.} & \textbf{Full SSL} & \textbf{MD + SSL} \\
			\midrule
			Data Exfiltration     & 0.280 & 0.511 & 0.272 & 0.651 & \textbf{0.699} \\
			Destructive Behaviors & 0.162 & 0.371 & 0.147 & 0.403 & \textbf{0.439} \\
			Privilege Escalation  & 0.364 & 0.381 & 0.381 & 0.348 & \textbf{0.455} \\
			Covert Execution      & 0.083 & 0.222 & 0.042 & 0.083 & \textbf{0.264} \\
			Resource Abuse        & 0.132 & 0.271 & 0.133 & 0.323 & \textbf{0.419} \\
			Credential Access     & 0.391 & 0.695 & 0.286 & 0.722 & \textbf{0.780} \\
			\midrule
			Macro F1              & 0.235 & 0.409 & 0.210 & 0.422 & \textbf{0.509} \\
			\bottomrule
		\end{tabular}
\caption{F1 scores on the Risk Assessment benchmark. Each column corresponds to a different input representation for the same fixed LLM judge.}
	\label{tab:security}
\end{table}

\begin{table}[h]
	\centering
	\small
	\begin{tabular}{lcccc}
		\toprule
		\textbf{Input} &\textbf{Macro Acc.} & \textbf{Macro Prec.} & \textbf{Macro Rec.} & \textbf{Macro F1} \\
		\midrule
		Desc Only & 0.714 & 0.823 & 0.148 & 0.235 \\
		Full MD & 0.765 & 0.828 & 0.283 & 0.409 \\
		SSL-Shallow &0.709 & 0.783 & 0.129 & 0.210 \\
		Full SSL & 0.778 & 0.841 & 0.315 & 0.422 \\
		MD + SSL & \textbf{0.801} & \textbf{0.884} & \textbf{0.382} & \textbf{0.509} \\
		\bottomrule
	\end{tabular}
\caption{Aggregate Risk Assessment results. Metrics are macro-averaged across the six dimensions.}
	\label{tab:security-aggregate}
\end{table}

The main result in Tables~\ref{tab:security} and~\ref{tab:security-aggregate} is that pairing SSL with the source document improves Risk Assessment over the complete source document alone. \texttt{Full SKILL.md + SSL} achieves the best macro F1, improving from 0.409 to 0.509 over \texttt{Full SKILL.md}, and also improves macro precision, recall, and accuracy. Full SSL alone is only slightly above the full-document baseline, suggesting that the main benefit comes from adding structured evidence to source context rather than replacing the source document.

The per-dimension results show that the combined view is strongest across all six dimensions, with the largest gains on data exfiltration, credential access, and resource abuse. These risks often require connecting scattered operational cues with source context; SSL surfaces the cues, while the original document helps interpret their risk relevance.

The recall gain indicates that the combined view recovers weak but concrete risk signals that text-only inputs often miss, without lowering precision. 
Appendix~\ref{app:bootstrap-ci} reports a paired bootstrap confidence interval for the main comparison: the macro-F1 improvement of \texttt{Full SKILL.md + SSL} over \texttt{Full SKILL.md} is 0.101, with a 95\% interval of [0.051, 0.152].

\section{Discussion}

\subsection{The Role of SSL as an Evidence Interface}

The experiments suggest that SSL is most effective as an evidence layer alongside the source document. It separates source-grounded signals so downstream systems can select task-relevant evidence without losing access to the original artifact.

This evidence-interface role connects the two evaluation settings: Skill Discovery mainly uses interface and workflow evidence; while Risk Assessment depends more on risk-relevant evidence about actions and resources. The tasks differ, but both benefit from making these evidence types explicit rather than leaving them entangled in long instructional prose.

\subsection{Why SSL Should Not Replace the Source Document}
SSL records evidence that can be typed and compared across skills, but the source document still carries richer context: examples, design rationale, safeguards, failure modes, and maintenance guidance. This division is acceptable for indexing and inspection, but it becomes important when a task requires judging quality, intent, or severity.

The experiments illustrate this boundary: In Skill Discovery, removing incidental prose can improve matching because invocation cues are less diluted; in Risk Assessment, the best result comes from combining the complete source document with SSL. The binary judge benefits from structured evidence about actions and resources, but the source text still helps interpret whether an operation is guarded, hypothetical, limited in scope, or embedded in human review. SSL is therefore best used with the source document: the structure points systems and reviewers to relevant evidence, while the source text supplies the context needed to interpret it.

\subsection{Implications for Skill-Centered Agent Systems}
The broader implication is that skill-centered systems need a shared manifest layer, not only better prompts over raw documentation. Without such a layer, it is difficult for registries, routers, policy checkers, and reviewers to avoid repeatedly recovering similar facts from the same \texttt{SKILL.md} file. SSL makes these facts persistent and source-adjacent, supporting registry indexing, phase inspection, and logic-level risk review while retaining access to the original document.

The evaluations in this paper focus on pre-execution skill management, where systems must select skills and assess risks before invocation. The same representation could also support skill construction and maintenance by making expected inputs, phase boundaries, dependencies, and resource effects explicit. During skill use, agents might also use SSL as an operational guide for choosing among candidate skills, tracking execution checkpoints, and recognizing steps that require confirmation or resource-sensitive handling.

\section{Conclusion}
We introduce SSL as a structured representation for agent skills, disentangling routing interfaces, execution structure, and low-level action/resource-use evidence from raw \texttt{SKILL.md} text. For evaluation, SSL-derived representations significantly outperform text-only baselines in the main evaluations: in Skill Discovery, \texttt{Desc + SSL-Rich} improves retrieval MRR@50 from 0.649 to 0.729 over the strongest non-SSL baseline; in Risk Assessment, the combined \texttt{SKILL.md} + SSL view improves macro F1 from 0.409 to 0.509 over full text alone.

To sum up, the representation is informative, but not sufficient by itself. In this paper, we mainly use SSL for skill management: enabling discovery of relevant skills, and supporting assessment of pre-execution risk. A natural next step is to move from managing skills to helping agents use them. Future work may refine SSL itself, for example by linking individual SSL graphs into repository-level skill graphs or enriching static normalization with runtime traces, and may study how agents might use SSL to select, compose, adapt, and reuse skills during task execution. We therefore view SSL not as a finished standard or a standalone security mechanism, but as a practical step toward more inspectable, reusable, and operationally actionable skill representations for agent systems.

\section{Limitations}
The study has several limitations at the current stage:
\begin{itemize}
	\item \textbf{Static runtime actions}: SSL is extracted from static artifacts and cannot determine dynamic actions by itself, such as downloaded payloads, constructed commands, or conditional resource access;
	\item \textbf{Normalizer fidelity}: SSL extraction depends on LLM-based normalization, which may omit relevant facts, over-regularize the source, or map ambiguous behavior into coarse enums for underspecified or obfuscated skills;
 	\item \textbf{Evaluation scope}: Our experiments cover Skill Discovery and Risk Assessment. They do not directly evaluate how SSL affects an agent's actual use of skills during planning, execution, monitoring, or post-hoc refinement;
	\item \textbf{About Skill Discovery benchmark}: The Skill Discovery benchmark uses automatically generated candidates filtered by a realism annotation protocol rather than fully human-authored requests.
	\item \textbf{Model-mediated Risk Assessment}: The current risk labels cover 252 completed skills and come from a multi-model binary voting pipeline with second-pass review for disagreed dimensions. The final evaluation uses a fixed LLM judge and model-generated gold labels, so the results should be interpreted as static risk-identification performance, not as a substitute for expert security audit or runtime safety evaluation.
\end{itemize}

\bibliographystyle{plainnat}
\bibliography{custom}

\appendix

\section{Definition of SSL Schema}
\label{app:ssl-schema}
This appendix gives the field-level realization of the SSL representation defined in Eq.~\ref{eq:ssl-representation}. We first state the graph-design principles behind the schema, then describe the scheduling, structural, and logical layers and their restricted vocabularies.

\subsection{Graph-Design Principles}
The SSL schema realizes Eq.~\ref{eq:ssl-representation} as a typed JSON graph with three linked representational levels. The Scheduling Layer is a single skill-level record that stores the interface information needed for selecting and invoking the skill. The Structural Layer is a directed scene graph whose nodes are execution phases and whose edges encode phase-level transitions. The Logical Layer is a directed logic-step graph whose nodes are source-grounded atomic actions and whose edges encode action-level control flow within a scene. Cross-level connectivity is deliberately restricted to containment relations and entry pointers: scenes belong to a skill, logic steps belong to scenes, and traversal begins from designated entry records. This restriction preserves clear boundaries among interface evidence, phase structure, and action/resource-use evidence while keeping cross-level relations explicit and limited.

The schema is designed to be \textit{compact}, \textit{grounded}, and \textit{comparable}. It records skill-level interface contracts and dependencies, scene-level phase structure, and logic-level data-flow cues, resource boundaries, and operational effects, while excluding open-ended attributes such as subjective skill quality, user persona, inferred developer intent, or speculative hidden behavior. Table~\ref{tab:ssl-schema} summarizes the principal fields for each layer, and Table~\ref{tab:ssl-vocabularies} lists the restricted vocabularies used to keep normalized records comparable. The following subsections explain how these fields and vocabularies are used.

\begin{table}[t]
	\centering
	\small
	\begin{tabularx}{\textwidth}{>{\raggedright\arraybackslash}p{0.16\textwidth}>{\raggedright\arraybackslash}p{0.25\textwidth}X}
		\toprule
		\textbf{Layer} & \textbf{Principal Fields} & \textbf{Purpose} \\
		\midrule
		Scheduling & \begin{tabular}[t]{@{}l@{}}
			\texttt{skill\_id} \\
			\texttt{skill\_name} \\
			\texttt{skill\_goal} \\
			\texttt{intent\_signature} \\
			\texttt{tags} \\
			\texttt{top\_pattern} \\
			\texttt{expected\_inputs} \\
			\texttt{expected\_outputs} \\
			\texttt{dependencies} \\
			\texttt{control\_flow\_features} \\
			\texttt{entry\_scene\_id} \\
			\texttt{subscenes}
		\end{tabular} & Exposes the skill's routing and invocation interface without requiring downstream systems to embed or parse the full source document. \\
    \addlinespace
    \midrule
		Structural & \begin{tabular}[t]{@{}l@{}}
			\texttt{scene\_id} \\
			\texttt{scene\_name} \\
			\texttt{scene\_type} \\
			\texttt{scene\_goal} \\
			\texttt{input} \\
			\texttt{output} \\
			\texttt{entry\_conditions} \\
			\texttt{exit\_conditions} \\
			\texttt{next\_scene\_rules} \\
			\texttt{entry\_logic\_step\_id} \\
			\texttt{contained\_logic\_steps}
		\end{tabular} & Represents the skill as a sequence or graph of typed execution phases, making multistep workflows inspectable. \\
    \addlinespace
    \midrule
		Logical & \begin{tabular}[t]{@{}l@{}}
			\texttt{logic\_step\_id} \\
			\texttt{act\_type} \\
			\texttt{actor} \\
			\texttt{object} \\
			\texttt{instrument} \\
			\texttt{input\_args} \\
			\texttt{output\_binding} \\
			\texttt{preconditions} \\
			\texttt{effects} \\
			\texttt{resource\_scope} \\
			\texttt{resource\_target} \\
			\texttt{next\_step\_rules}
		\end{tabular} & Records bounded action and resource-use facts that are useful for data-flow inspection and Risk Assessment. \\
		\bottomrule
	\end{tabularx}
\caption{Core fields of the SSL schema.}
	\label{tab:ssl-schema}
\end{table}

\begin{table}[t]
	\centering
	\small
	\begin{tabularx}{\textwidth}{lX}
		\toprule
		\textbf{Field or Constraint} & \textbf{Allowed Values} \\
		\midrule
		\texttt{scene\_type} & \texttt{PREPARE}, \texttt{ACQUIRE}, \texttt{REASON}, \texttt{ACT}, \texttt{VERIFY}, \texttt{RECOVER}, \texttt{FINALIZE} \\
    \addlinespace
		\texttt{act\_type} & \texttt{READ}, \texttt{SELECT}, \texttt{COMPARE}, \texttt{VALIDATE}, \texttt{INFER}, \texttt{WRITE}, \texttt{UPDATE\_STATE}, \texttt{CALL\_TOOL}, \texttt{REQUEST}, \texttt{TRANSFER}, \texttt{NOTIFY}, \texttt{TERMINATE} \\
    \addlinespace
		\texttt{resource\_scope} & \texttt{MEMORY}, \texttt{LOCAL\_FS}, \texttt{CODEBASE}, \texttt{PROCESS}, \texttt{USER\_DATA}, \texttt{CREDENTIALS}, \texttt{NETWORK}, \texttt{OTHER} \\
    \addlinespace
		\texttt{target} terminal values & \texttt{END\_SUCCESS}, \texttt{END\_FAIL}, \texttt{YIELD\_SUCCESS}, \texttt{YIELD\_FAIL} \\
		\bottomrule
	\end{tabularx}
	\caption{Restricted vocabularies used by the SSL normalizer.}
	\label{tab:ssl-vocabularies}
\end{table}

\subsection{Scheduling Layer}
The scheduling layer stores the skill-level interface. Its role is to expose what a skill is for and how it can be invoked without requiring downstream systems to re-embed or re-parse the entire source document. Fields such as \texttt{skill\_goal}, \texttt{intent\_signature}, \texttt{tags}, \texttt{top\_pattern}, \texttt{expected\_inputs}, \texttt{expected\_outputs}, and \texttt{dependencies} capture the stable capability-facing surface of the skill, while \texttt{entry\_scene\_id} and \texttt{subscenes} connect that surface to the execution graphs below. To avoid duplicating the full execution graph at the scheduling level, the top-level \texttt{control\_flow\_features} field remains intentionally coarse, storing summary signals such as whether the skill contains branching, loops, tool calls, or sensitive-resource access.

\subsection{Structural Layer}
The structural layer represents a skill as a scene-level execution graph. Instead of an arbitrary text span or sentence block, a scene should denote a coherent execution phase with its own goal, data contract, and exit conditions. Accordingly, fields such as \texttt{scene\_type}, \texttt{scene\_goal}, \texttt{input}, \texttt{output}, \texttt{entry\_conditions}, \texttt{exit\_conditions}, and \texttt{next\_scene\_rules} describe how the skill unfolds at the phase level. Control flow is represented here through \texttt{next\_scene\_rules}, whose nested \texttt{target} field must either name another scene in the same graph or use a reserved terminal value. \texttt{END\_SUCCESS} and \texttt{END\_FAIL} therefore act as closed control-flow outcomes for the scene graph, not free-form labels; together with the logic-level \texttt{YIELD\_SUCCESS} and \texttt{YIELD\_FAIL} values, they form the terminal-target vocabulary listed in Table~\ref{tab:ssl-vocabularies}. The \texttt{scene\_type} field is likewise drawn from a closed inventory so that phase categories remain comparable across skills; as listed in Table~\ref{tab:ssl-vocabularies}, these types are \texttt{PREPARE}, \texttt{ACQUIRE}, \texttt{REASON}, \texttt{ACT}, \texttt{VERIFY}, \texttt{RECOVER}, and \texttt{FINALIZE}.

\subsection{Logical Layer}
The logical layer represents the operations that implement each scene. It is defined over source-grounded atomic actions, where ``atomic'' means the smallest operational unit that the source artifact supports without inventing missing implementation detail. A logic step should usually be split when the source supports a change in action type, resource boundary, effect, or control-flow outcome; for example, it is preferable to represent reading a local file, calling an external tool, and writing back to the codebase as distinct steps when those operations are separately recoverable from the artifact. Fields such as \texttt{act\_type}, \texttt{actor}, \texttt{object}, \texttt{instrument}, \texttt{input\_args}, \texttt{output\_binding}, \texttt{preconditions}, \texttt{effects}, \texttt{resource\_scope}, \texttt{resource\_target}, and \texttt{next\_step\_rules} describe this action/resource-use evidence. Data-flow variables start with a prefix \texttt{\$}, while unprefixed strings denote literal labels, resources, or natural-language conditions. Within a scene, \texttt{next\_step\_rules} route among logic steps; their targets must either name another in-scope step or use \texttt{YIELD\_SUCCESS} or \texttt{YIELD\_FAIL} to return control to the enclosing scene.

As listed in Table~\ref{tab:ssl-vocabularies}, the logical layer uses two closed vocabularies. The \texttt{act\_type} field records the action primitive: \texttt{READ} for information access; \texttt{SELECT}, \texttt{COMPARE}, \texttt{VALIDATE}, and \texttt{INFER} for decision and transformation steps; \texttt{WRITE} and \texttt{UPDATE\_STATE} for mutation; \texttt{CALL\_TOOL}, \texttt{REQUEST}, \texttt{TRANSFER}, and \texttt{NOTIFY} for external interaction or data movement; and \texttt{TERMINATE} for explicit control termination. The \texttt{resource\_scope} field separately records the operational boundary touched by the action. Both vocabularies are intentionally coarse, preventing idiosyncratic labels while preserving distinctions useful for comparing execution behavior, resource contact, and risk-relevant operations.

\section{A Complete Example of SSL}
\label{app:ssl-example}
The following compact instance illustrates what a normalized SSL object looks like after annotation.

{\scriptsize
\begin{verbatim}
{
  "skill": {
    "skill_id": "SKILL_WRITING_REFINER",
    "skill_name": "Writing Refiner",
    "skill_goal": "Revise user-provided text for clarity and concision.",
    "top_pattern": "GUIDE_AND_APPLY",
    "expected_inputs": [
      {"name": "draft_text", "type": "str"},
      {"name": "writing_context", "type": "str"}
    ],
    "expected_outputs": [
      {"name": "revised_text", "type": "str"},
      {"name": "applied_principles", "type": "list"}
    ],
    "dependencies": [
      {"type": "permission", "value": "filesystem.read"},
      {"type": "capability", "value": "text_processing"}
    ],
    "tags": ["writing", "editing", "documentation"],
    "intent_signature": ["improve this text", "edit for concision"],
    "control_flow_features": {
      "has_branch": true,
      "has_loop": false,
      "has_tool_call": true,
      "touches_sensitive_resources": false
    },
    "entry_scene_id": "S_PREPARE",
    "subscenes": ["S_PREPARE", "S_ACQUIRE", "S_REVISE"]
  },
  "scenes": [
    {
      "scene_id": "S_PREPARE",
      "scene_name": "Prepare request",
      "scene_type": "PREPARE",
      "scene_goal": "Validate the request and infer the editing intent.",
      "input": ["$draft_text", "$writing_context"],
      "output": ["$parsed_intent", "$target_principles"],
      "entry_conditions": ["skill_dispatched"],
      "exit_conditions": ["writing_task_clarified"],
      "entry_logic_step_id": "L_VALIDATE_INPUT",
      "contained_logic_steps": ["L_VALIDATE_INPUT", "L_PARSE_CONTEXT"],
      "next_scene_rules": [
        {"condition": "success", "target": "S_ACQUIRE"},
        {"condition": "default", "target": "END_FAIL"}
      ]
    },
    {
      "scene_id": "S_ACQUIRE",
      "scene_name": "Acquire style guidance",
      "scene_type": "ACQUIRE",
      "scene_goal": "Load the style guidance needed for the task.",
      "input": ["$writing_context", "$target_principles"],
      "output": ["$loaded_guidelines"],
      "entry_conditions": ["writing_task_clarified"],
      "exit_conditions": ["guidelines_loaded"],
      "entry_logic_step_id": "L_SELECT_GUIDE",
      "contained_logic_steps": ["L_SELECT_GUIDE", "L_READ_GUIDE"],
      "next_scene_rules": [
        {"condition": "success", "target": "S_REVISE"},
        {"condition": "default", "target": "END_FAIL"}
      ]
    },
    {
      "scene_id": "S_REVISE",
      "scene_name": "Revise draft",
      "scene_type": "REASON",
      "scene_goal": "Apply the selected rules and return the revision.",
      "input": ["$draft_text", "$loaded_guidelines", "$parsed_intent"],
      "output": ["$revised_text", "$applied_principles"],
      "entry_conditions": ["guidelines_loaded"],
      "exit_conditions": ["text_revised", "summary_generated"],
      "entry_logic_step_id": "L_PARSE_GUIDE",
      "contained_logic_steps": [
        "L_PARSE_GUIDE", "L_SELECT_RULES", "L_APPLY_EDITING"
      ],
      "next_scene_rules": [
        {"condition": "success", "target": "END_SUCCESS"},
        {"condition": "default", "target": "END_FAIL"}
      ]
    }
  ],
  "logic_steps": [
    {
      "logic_step_id": "L_VALIDATE_INPUT",
      "act_type": "VALIDATE",
      "input_args": ["$draft_text", "$writing_context"],
      "output_binding": "$input_valid",
      "actor": "skill",
      "object": "user_input",
      "instrument": null,
      "preconditions": ["skill_dispatched"],
      "effects": ["$input_valid == true"],
      "resource_scope": "MEMORY",
      "resource_target": "working_memory.user_request",
      "next_step_rules": [
        {"condition": "$input_valid == true", "target": "L_PARSE_CONTEXT"},
        {"condition": "default", "target": "YIELD_FAIL"}
      ]
    },
    {
      "logic_step_id": "L_PARSE_CONTEXT",
      "act_type": "INFER",
      "input_args": ["$writing_context"],
      "output_binding": "$target_principles",
      "actor": "skill",
      "object": "writing_context",
      "instrument": null,
      "preconditions": ["$input_valid == true"],
      "effects": ["writing_task_clarified"],
      "resource_scope": "MEMORY",
      "resource_target": "working_memory",
      "next_step_rules": [
        {"condition": "always", "target": "YIELD_SUCCESS"}
      ]
    },
    {
      "logic_step_id": "L_SELECT_GUIDE",
      "act_type": "SELECT",
      "input_args": ["$writing_context"],
      "output_binding": "$guide_file_path",
      "actor": "skill",
      "object": "guide_repository",
      "instrument": null,
      "preconditions": ["writing_task_clarified"],
      "effects": ["primary_guide_selected"],
      "resource_scope": "MEMORY",
      "resource_target": "guide_index",
      "next_step_rules": [
        {"condition": "always", "target": "L_READ_GUIDE"}
      ]
    },
    {
      "logic_step_id": "L_READ_GUIDE",
      "act_type": "READ",
      "input_args": ["$guide_file_path"],
      "output_binding": "$loaded_guidelines",
      "actor": "skill",
      "object": "style_guide_file",
      "instrument": "filesystem.read",
      "preconditions": ["primary_guide_selected"],
      "effects": ["guidelines_loaded"],
      "resource_scope": "LOCAL_FS",
      "resource_target": "$guide_file_path",
      "next_step_rules": [
        {"condition": "success", "target": "YIELD_SUCCESS"},
        {"condition": "default", "target": "YIELD_FAIL"}
      ]
    },
    {
      "logic_step_id": "L_PARSE_GUIDE",
      "act_type": "INFER",
      "input_args": ["$loaded_guidelines"],
      "output_binding": "$parsed_rules",
      "actor": "skill",
      "object": "guidelines",
      "instrument": null,
      "preconditions": ["guidelines_loaded"],
      "effects": ["rules_available"],
      "resource_scope": "MEMORY",
      "resource_target": "working_memory",
      "next_step_rules": [
        {"condition": "always", "target": "L_SELECT_RULES"}
      ]
    },
    {
      "logic_step_id": "L_SELECT_RULES",
      "act_type": "SELECT",
      "input_args": ["$parsed_rules", "$target_principles"],
      "output_binding": "$selected_rules",
      "actor": "skill",
      "object": "rule_set",
      "instrument": null,
      "preconditions": ["rules_available"],
      "effects": ["relevant_rules_selected"],
      "resource_scope": "MEMORY",
      "resource_target": "working_memory",
      "next_step_rules": [
        {"condition": "always", "target": "L_APPLY_EDITING"}
      ]
    },
    {
      "logic_step_id": "L_APPLY_EDITING",
      "act_type": "INFER",
      "input_args": ["$draft_text", "$selected_rules"],
      "output_binding": "$revised_text",
      "actor": "skill",
      "object": "draft_text",
      "instrument": "text_processing",
      "preconditions": ["relevant_rules_selected"],
      "effects": ["text_revised", "$applied_principles generated"],
      "resource_scope": "MEMORY",
      "resource_target": "working_memory.draft_text",
      "next_step_rules": [
        {"condition": "success", "target": "YIELD_SUCCESS"},
        {"condition": "default", "target": "YIELD_FAIL"}
      ]
    }
  ]
}
\end{verbatim}
}

\section{Prompting Protocol of the SSL Normalizer}
\label{app:normalizer-prompt}
The SSL Normalizer employs a constrained \texttt{NL2JSON} instruction with four steps. The normalizer backend is DeepSeek-V3.2~\citep{liu2025deepseek}. The 6,184 released SSL records are produced by an initial DeepSeek-V3.2-Thinking pass followed by a DeepSeek-V3.2-Nothinking retry pass for remaining or malformed records, both at temperature 0.1. Matching the four-pass pipeline in Section~\ref{sec:methodology}, the implementation prompt is organized around the stages as shown in Table~\ref{tab:normalizer-prompt}. It also includes a minimal one-shot example of a skill that validates inputs, calls a remote API, yields success or failure from the logic-step graph, and then terminates the scene graph. The paragraphs below describe five cross-cutting implementation details that apply throughout normalization, \textit{model and decoding parameters}, \textit{grounded output}, \textit{validation}, \textit{retry behavior}, and \textit{normalization yield}:

\paragraph{Model and decoding parameters.}
The initial normalization pass uses DeepSeek-V3.2 API with thinking enabled and temperature 0.1. Remaining failures are regenerated with DeepSeek-V3.2 without thinking, also at temperature 0.1. All accepted records are produced by inference-only calls followed by deterministic JSON parsing and validation;

\paragraph{Grounded output.}
The normalizer is instructed to populate the fixed schema only with evidence grounded in the source artifact. It may reorganize source evidence into typed fields, but it is not allowed to infer hidden intent, add unstated runtime behavior, or complete missing execution steps from general background knowledge. The prompt also constrains the output channel: the model must return raw JSON only, without Markdown fences, prose explanations, comments, or conversational prefixes, and it must use the closed vocabularies defined by the SSL schema;

\paragraph{Validation.}
We separate hard structural validation from softer semantic checks: Hard validation requires parseable JSON, all required top-level fields, globally unique identifiers, valid enum values, valid containment links, valid entry pointers, and transition targets that either name an in-scope node or use a reserved terminal symbol; Softer checks include whether scene-level outputs are supported by logic-step bindings and whether data-flow references are internally consistent. Because source skill documents often describe data flow only partially, these softer checks serve as tools for repair and quality control instead of strict rejection criteria;

\paragraph{Retry behavior.}
Outputs that fail parsing or hard validation are rejected and regenerated. The initial normalizer allows up to five API attempts per skill; the retry pass allows up to three attempts for remaining failures. If repeated attempts cannot ground a field in the source, the normalizer does not invent a value; it leaves the field empty, \texttt{null}, or at the coarsest supported category. This makes normalization conservative: SSL exposes what the artifact says or strongly implies, but it does not claim to reveal hidden runtime behavior;

\paragraph{Normalization yield.}
The raw collection contains 6,300 skill directories with \texttt{SKILL.md} files, of which 6,184 produce valid SSL records after bounded validation and retry, giving a final normalization yield of 98.16\%. The remaining 116 skills are excluded because they do not produce parseable, schema-valid, and source-grounded records under the retry budget. Invalid outputs are rejected rather than silently repaired or partially accepted, so the released corpus reflects a bounded, high-yield normalization pipeline rather than an unvalidated filtering step;

\begin{table}[t]
	\centering
	\small
	\begin{tabularx}{\textwidth}{p{0.24\textwidth}X}
		\toprule
		\textbf{Pipeline Stage} & \textbf{Prompt Constraint} \\
		\midrule
		Pass 1: Skill record extraction & Extract the scheduling record: \texttt{skill\_goal}, \texttt{top\_pattern}, \texttt{intent\_\space signature}, \texttt{expected\_inputs}, \texttt{expected\_outputs}, \texttt{dependencies}, \texttt{tags}, \texttt{control\_flow\_features}, \texttt{entry\_scene\_id}, and \texttt{subscenes}; \\
    \addlinespace
		Pass 2: Scene decomposition & Decompose the skill into two to five macro-level scenes when supported by the source; keep scene records at the phase/milestone level, with typed scene categories, data contracts, entry and exit conditions, and \texttt{next\_scene\_rules}; \\
    \addlinespace
		Pass 3: Logic-step expansion & Expand each scene into grounded atomic operations; assign allowed \texttt{act\_type} values, roles, instruments, \texttt{resource\_scope}, \texttt{resource\_target}, \texttt{input\_args}, \texttt{output\_binding}, \texttt{preconditions}, \texttt{effects}, and \texttt{next\_step\_rules}; \\
    \addlinespace
		Pass 4: Verification and validation & Verify layer alignment and reject malformed outputs: enforce globally unique identifiers, valid enums, valid transition targets, valid containment links, valid entry pointers, and scene outputs backed by logic-step bindings. Invalid outputs are strictly retried, never silently accepted; \\
    \addlinespace
		Output mode & Act only as an \texttt{NL2JSON} converter for skill artifacts; output raw JSON only, with no Markdown fences, prose explanation, or conversational text. \\
		\bottomrule
	\end{tabularx}
	\caption{Prompt stages corresponding to the four-pass SSL Normalizer pipeline.}
	\label{tab:normalizer-prompt}
\end{table}

\section{Human Audit of Fidelity of SSL Normalizer}
\label{app:normalizer-human-audit}
To quantify the extraction fidelity of the LLM-based normalizer, we conduct a lightweight human audit on 100 sampled skills. The audit is designed as a binary source-grounding check rather than a full manual reconstruction of SSL. For each sampled skill, annotators inspect the original \texttt{SKILL.md} together with selected SSL claims produced by the normalizer and answer a single question: whether the SSL claim is supported by the source artifact.

\paragraph{Annotation guideline.}
Each audit item is labeled \textit{Yes} if the claim is explicitly stated in the source document or follows from a direct, low-ambiguity paraphrase of the source. For example, an instruction to ``write the final report to the project directory'' can support a logical \texttt{WRITE} action over \texttt{LOCAL\_FS} or \texttt{CODEBASE}. An item is labeled \textit{No} if the claim is not supported by the source, adds unstated tools or resources, treats an optional or hypothetical behavior as mandatory, assigns an incompatible scene or action type, or infers hidden runtime intent beyond the artifact. Table~\ref{tab:normalizer-audit-rubric} summarizes the audit rubric.

\begin{table}[h]
	\centering
	\small
	\begin{tabularx}{\textwidth}{p{0.1\textwidth}p{0.35\textwidth}X}
		\toprule
		\textbf{Layer} & \textbf{Audited Claims} & \textbf{Binary Decision Rule} \\
		\midrule
		Scheduling & Skill goal, tags, expected inputs and outputs, dependencies, resource scopes & Mark \textit{Yes} when the interface-level claim is stated or directly implied by the source; mark \textit{No} when it adds an unsupported capability, dependency, input, output, or resource. \\
    \addlinespace
		Structural & Scene type, scene goal, phase boundary, scene transition & Mark \textit{Yes} when the phase abstraction and transition are recoverable from the documented workflow; mark \textit{No} when the normalizer invents a phase, misses the role of a phase, or turns loose prose into unsupported control flow. \\
    \addlinespace
		Logical & Act type, tool call, read/write/transfer effect, resource boundary & Mark \textit{Yes} when the action and resource-use claim is grounded in an instruction, dependency, or described side effect; mark \textit{No} when it infers an unstated operation or overstates runtime behavior. \\
		\bottomrule
	\end{tabularx}
	\caption{Binary human-audit rubric for checking whether SSL claims are supported by the original skill artifact.}
	\label{tab:normalizer-audit-rubric}
\end{table}

\paragraph{Metric and result.}
We report support accuracy, defined as the fraction of audited SSL claims labeled \textit{Yes}:
\[
\text{Support Accuracy} =
\frac{\#\{\text{audited claims labeled Yes}\}}{\#\{\text{all audited claims}\}}.
\]
On the 100-skill audit set, the normalizer achieves an item-level support accuracy of 83\%. This result indicates that most audited SSL fields are grounded in the source artifacts, while leaving room for improvement in cases where the normalizer over-abstracts weakly specified workflow steps or resource effects.

\section{Construction and Quality Control of Skill Discovery Benchmark}
\label{app:retrieval-by-type}
The Skill Discovery benchmark is built from our collected and formalized corpus of 6,184 skills. We sample 200 source skills and build a candidate pool of automatically generated requests from two complementary views. The first generator receives normalized source-skill evidence such as the skill goal, tags, scene types, and control-flow features; the second receives only the raw \texttt{SKILL.md} file. Both generators use DeepSeek-V3.2 at temperature 0.7 and are instructed to produce natural user requests without directly naming the source skill. After deduplication, the combined candidate pool contains 806 requests.

We then run an independent realism annotation pass with DeepSeek-V3.2 at temperature 0.0. This pass does not see SSL fields or retrieval outputs: the annotator receives only the candidate query and the raw \texttt{SKILL.md} of the labeled source skill. It rates realism, source grounding, and document proximity on a 1--5 scale, where higher realism and grounding are better but higher document proximity means the query is too close to source wording or implementation details; it also marks whether the query leaks the exact skill name. A query is kept only if it has realism score at least 4, grounding score at least 4, document-proximity score at most 2, no name leakage, and length at most 120 characters. After automatic filtering, we manually audit a random sample of generated queries. A query passes the audit if it is answerable by the source skill, matches its intended query style, and does not reveal the skill name. More than 95\% of audited queries pass these checks; failed cases are removed or regenerated before the final query set is fixed. This yields 431 intent-level queries: 95 \textit{functional}, 86 \textit{constraint-based}, 88 \textit{compositional}, 73 \textit{safety-oriented}, and 89 \textit{scenario-style} requests.

Because many skills in the corpus have overlapping names or capabilities, the benchmark treats the source skill as the single labeled relevant item. This strict protocol counts near-equivalent neighboring skills as errors, which may understate retrieval quality but keeps the metric definition unambiguous. All settings use the same embedding model, FAISS ranking pipeline, and input representations as shown in Table~\ref{tab:retrieval}.

The source-outline controls in Table~\ref{tab:retrieval} are constructed without SSL fields. They deterministically extract headings, short bullet items, interface- or resource-looking lines, and short prose spans from the original \texttt{SKILL.md}. This provides a deterministic non-SSL document-enrichment control for testing whether the SSL improvement can be explained by generic source-text expansion alone. Table~\ref{tab:retrieval-by-type} reports MRR@50 by query type.

\begin{table}[h]
	\centering
	\scriptsize
	\begin{tabularx}{\textwidth}{p{0.23\textwidth}Xcccccc}
		\toprule
		\textbf{Group} & \textbf{Input} & \textbf{Overall} & \textbf{Func.} & \textbf{Const.} & \textbf{Comp.} & \textbf{Safety} & \textbf{Scenario} \\
		\midrule
		\textit{Non-SSL baselines} & Desc Only & 0.588 & 0.509 & 0.555 & 0.599 & 0.548 & 0.727 \\
		& Full \texttt{SKILL.md} & 0.645 & 0.600 & 0.662 & 0.666 & 0.530 & 0.748 \\
		& Source Outline & 0.592 & 0.573 & 0.618 & 0.630 & 0.470 & 0.646 \\
		& Desc + Source Outline & 0.649 & 0.615 & 0.678 & 0.695 & 0.539 & 0.701 \\
		\midrule
		\textit{Description + SSL variants} & Desc + SSL-Shallow & 0.716 & \textbf{0.699} & 0.711 & 0.732 & \textbf{0.700} & 0.734 \\
		& Desc + SSL-Sched & 0.694 & 0.663 & 0.711 & 0.710 & 0.643 & 0.736 \\
		& Desc + SSL-Rich & \textbf{0.729} & 0.695 & \textbf{0.731} & \textbf{0.778} & 0.674 & 0.762 \\
		\midrule
		\textit{Full-document + SSL variants} & Full \texttt{SKILL.md} + SSL-Shallow & 0.664 & 0.638 & 0.676 & 0.709 & 0.499 & 0.772 \\
		& Full \texttt{SKILL.md} + SSL-Sched & 0.676 & 0.632 & 0.705 & 0.733 & 0.503 & 0.782 \\
		& Full \texttt{SKILL.md} + SSL-Rich & 0.681 & 0.631 & 0.706 & 0.725 & 0.528 & \textbf{0.793} \\
		\bottomrule
	\end{tabularx}
	\caption{MRR@50 by query type on the 431-query intent-level retrieval benchmark.}
	\label{tab:retrieval-by-type}
\end{table}

\section{Construction and Rubric of Risk Assessment Benchmark}
\label{app:risk-assessment-benchmark}
\label{app:risk-dimensions}
Section~\ref{sec:evaluation} describes the sampling strategy, gold-label construction, evaluator model, and input representations for the Risk Assessment benchmark. This appendix specifies the quality-control protocol, binary aggregation rule, and boundary rules for the six risk dimensions, summarized in Table~\ref{tab:risk-dimensions}.

The 252-skill benchmark is drawn from the same 6,184-skill corpus used in Skill Discovery. Sampling is stratified by coarse risk-relevant signals derived from normalized metadata. Skills with tool calls plus network or credential resources are assigned to a high-signal stratum, skills with branching or loops to a medium-signal stratum, and the remaining skills to a low-signal stratum. The stratification is used only to make the evaluation set contain enough observable risk-relevant evidence for discrimination; it is not used as a target label, and the final gold labels are assigned independently by the labeling pipeline.

Gold labels are produced by three high-capability models: Gemini-3.1-pro-preview, Claude-Sonnet-4.5, and GPT-5. Each model receives the complete source \texttt{SKILL.md} and the complete SSL record. The SSL record serves as an inspection aid for finding candidate evidence, while the target label is defined over risk signals observable in the skill artifact. For each skill and dimension, the model outputs only \textit{risk} or \textit{no risk} plus a short rationale. Outputs must parse into the six-dimension schema; malformed or incomplete outputs are retried, and checkpointed records are re-parsed before aggregation.

The first-round label is the majority vote among the three models. If all three models agree, that label is accepted directly. If a dimension receives both \textit{risk} and \textit{no risk} votes, we run a second review pass for that dimension only. The review prompt receives the source document, the SSL record, and the first-round labels and rationales; the same three models then re-label only the disputed dimensions, and the final label is the review-round majority vote. Table~\ref{tab:risk-gold-distribution} reports the resulting positive-label counts and the number of dimensions that required second-pass review.

We also run a source-grounded manual audit on a sampled subset of final labels. Auditors check the original \texttt{SKILL.md} against the rubric and the model rationale, and a positive label is accepted only when the source artifact visibly supports the claimed risk signal. Evidence that appears only as an SSL field is not sufficient unless the corresponding source evidence can be located in the original artifact.

As for the six dimensions, they are labeled independently rather than mutually exclusive classes: for example, a skill that reads a credential and sends it to a remote endpoint may trigger both credential-access and data-exfiltration risk.

\begin{table}[h]
	\centering
	\small
	\begin{tabular}{lccc}
		\toprule
		\textbf{Dimension} & \textbf{Risk Labels} & \textbf{Positive Rate} & \textbf{Reviewed Disagreements} \\
		\midrule
		Data exfiltration & 123 / 252 & 0.488 & 81 \\
		Destructive behavior & 99 / 252 & 0.393 & 69 \\
		Privilege escalation & 15 / 252 & 0.060 & 39 \\
		Covert execution & 46 / 252 & 0.183 & 62 \\
		Resource abuse & 98 / 252 & 0.389 & 107 \\
		Credential access & 106 / 252 & 0.421 & 44 \\
		\bottomrule
	\end{tabular}
	\caption{Binary gold-label distribution for the 252-skill Risk Assessment benchmark. Reviewed disagreements count dimensions whose first-round three-model labels were not unanimous and were therefore sent to the second review pass.}
	\label{tab:risk-gold-distribution}
\end{table}

\begin{table}[h]
	\centering
	\small
	\begin{tabularx}{\textwidth}{p{0.18\textwidth}XX}
		\toprule
		\textbf{Dimension} & \textbf{Positive signal} & \textbf{Boundary with other dimensions} \\
		\midrule
		Data exfiltration & Sends, uploads, or otherwise exposes local, user, workspace, or third-party service data to an external sink. & Requires outward data movement or disclosure risk; simply reading data is not enough unless the flow can leave the local/user-controlled context. \\
    \addlinespace
		Destructive behavior & Deletes, overwrites, corrupts, disables, revokes, or irreversibly mutates files, databases, cloud resources, accounts, or code assets. & Captures asset damage or irreversible state change; expensive execution without asset damage is scored under resource abuse. \\
    \addlinespace
		Privilege escalation & Obtains, requests, or exercises authority beyond the least privilege needed for the stated task, including admin scopes, bypassed authorization, or stronger execution permissions. & Focuses on authority expansion; ordinary use of an explicitly required credential is not escalation by itself. \\
    \addlinespace
		Covert execution & Runs in a hidden, background, persistent, opaque, or hard-to-audit manner, or hides material actions from the user. & Captures lack of observability; normal automation is not covert when actions, triggers, and outputs are explicit. \\
    \addlinespace
		Resource abuse & Enables runaway loops, large fan-out calls, scraping bursts, unbounded retries, excessive compute, API-cost amplification, or abusive automation. & Captures consumption and scale; data theft or asset damage should also be scored under the corresponding data or destructive dimension. \\
    \addlinespace
		Credential access & Reads, collects, stores, logs, transmits, or exposes secrets, tokens, API keys, passwords, private keys, cookies, or credential files. & Focuses on access to secrets; if those secrets are also sent outward, data exfiltration should be scored separately. \\
		\bottomrule
	\end{tabularx}
	\caption{Dimensions and boundary rules for Risk Assessment.}
	\label{tab:risk-dimensions}
\end{table}

\section{Reproducibility, Release, and Responsible Use}
\label{app:reproducibility-release}
All reported experiments are inference-only: reproducibility depends on hosted-model inference, embedding generation, ranking, and aggregation rather than model training or fine-tuning. The raw collection contains 6,300 skill directories with \texttt{SKILL.md} files, of which 6,184 have valid normalized SSL records, giving a final normalization yield of 98.16\%.

The main reproducibility cost is LLM inference, while local computation is comparatively lightweight. Rebuilding the reported artifacts requires inference calls for SSL normalization, query construction, realism annotation, Risk Assessment gold construction, and fixed-judge evaluation. In our run, this corresponds to about $6.3$k first-pass normalizer calls, roughly 150 second-pass calls for initially failed normalizations, 400 query-generation calls, 806 realism-annotation calls, at least 756 first-round Risk Assessment gold-construction calls, additional review calls for disputed dimensions, and 1,260 fixed-judge Risk Assessment calls. The local retrieval pipeline embeds 62,271 text inputs and builds FAISS inner-product indexes over L2-normalized Qwen3-Embedding-0.6B vectors; stored embeddings for one representation occupy about 25 MB. FAISS ranking, metric computation, JSON parsing, majority-vote aggregation, and 20,000 paired bootstrap resamples run on commodity CPU memory. Wall-clock time is dominated by hosted-model latency, model availability, and provider-side concurrency limits; we do not report a fixed dollar cost because API pricing and batching policies change over time.

The planned release includes the SSL schema and prompting guidelines, normalized SSL records, Skill Discovery queries and labels, Risk Assessment rubrics and labels, and scripts for rebuilding the reported tables from released artifacts. The submission version does not provide a public code-and-data package; the release package will be made available upon acceptance, as stated in the abstract footnote. To respect existing assets, each released skill artifact or derived record will retain source attribution and available license metadata. When a source license is unavailable or does not permit redistribution of the raw artifact, the release will provide derived annotations and source pointers rather than republishing the original text.

The work is intended to support more inspectable skill management and pre-execution review. The released risk labels describe artifact-level evidence and should not be interpreted as executable attack instructions or as guarantees of runtime safety. Before release, artifacts are checked for obvious private credentials or non-public personal data, and unsuitable entries are excluded or redacted. The manual audits reported in the paper are author quality-control checks over artifacts and model outputs; they are not crowdsourcing experiments or human-subject studies.

\section{Bootstrap Confidence Intervals for Main Results}
\label{app:bootstrap-ci}
We report paired non-parametric bootstrap confidence intervals for the two main empirical comparisons in Section~\ref{sec:evaluation}. For Skill Discovery, the resampling unit is the query: each bootstrap sample draws 431 queries with replacement and recomputes MRR@50. For Risk Assessment, the resampling unit is the skill: each bootstrap sample draws 252 skills with replacement and recomputes the six-dimension binary macro F1. We use 20,000 bootstrap replicates with a fixed random seed, and Table~\ref{tab:bootstrap-ci} reports percentile 95\% confidence intervals for the paired improvement.

\begin{table}[h]
	\centering
	\footnotesize
	\begin{tabularx}{\textwidth}{p{0.14\textwidth}p{0.1\textwidth}Xccc}
		\toprule
		\textbf{Task} & \textbf{Metric} & \textbf{Comparison} & \textbf{Base} & \textbf{SSL} & \textbf{$\Delta$ 95\% CI} \\
		\midrule
		Skill Discovery & MRR@50 & Desc + Source Outline vs. Desc + SSL-Rich & 0.649 & 0.729 & [0.051, 0.111] \\
    \addlinespace
		Risk Assessment & Macro F1 & Full \texttt{SKILL.md} vs. Full \texttt{SKILL.md} + SSL & 0.409 & 0.509 & [0.051, 0.152] \\
		\bottomrule
	\end{tabularx}
	\caption{Paired bootstrap confidence intervals for the main improvements reported in Section~\ref{sec:evaluation}. The intervals are computed over the paired difference between the SSL-augmented input and its corresponding text-only baseline.}
	\label{tab:bootstrap-ci}
\end{table}

\section{Case Studies and Qualitative Analysis}
\label{app:case-studies}
We inspect per-example outputs to make the aggregate trends in Section~\ref{sec:evaluation} concrete. The first two cases are selected from large positive deltas in the corresponding evaluation outputs, and the third is selected from a counterexample where adding SSL worsens the risk judgment. For binary risk labels, we report dimensions in the order: \textit{data exfiltration, destructive behavior, privilege escalation, covert execution, resource abuse,} and \textit{credential access}. The following cases illustrate these patterns more concretely:

\textbf{Case 1: Interface signals recover a missed retrieval target.} For the query ``create an Excel workbook that automatically updates financial data and applies formatting,'' the relevant skill is \texttt{xlsx-official}. \texttt{Desc Only} ranks the source skill at 2,493 among 6,184 candidates, while \texttt{Desc + SSL-Rich} ranks it first; the full-document variants rank it 12th without SSL and 6th with SSL-Rich. The improvement is consistent with the normalized fields: the skill exposes tags such as \texttt{excel}, \texttt{xlsx}, \texttt{financial-modeling}, \texttt{formatting}, and \texttt{automation}; its scene profile contains preparation, action, and verification stages; and its resource scopes include \texttt{LOCAL\_FS} and \texttt{PROCESS}. These interface-level signals align directly with the query's required artifact type, domain, and workflow, whereas the raw description alone is too short to reliably distinguish it from many generic data-analysis or spreadsheet-adjacent skills;

\textbf{Case 2: Risk-relevant evidence changes a risk judgment.} For \texttt{gws-gmail}, the gold labels are $(\textit{risk},\textit{risk},\textit{no},\textit{no},\textit{risk},\textit{no})$. With only the full \texttt{SKILL.md}, DeepSeek-V3.2 predicts only data-exfiltration risk and misses destructive and resource-abuse signals. With \texttt{SKILL.md} + SSL, it matches all six labels. The source document lists Gmail operations such as sending, replying, forwarding, watching, and managing messages, but these capabilities are spread across helper commands and API-resource descriptions. The SSL view makes the relevant operations and resource contact explicit, helping the judge connect email sending and mailbox management to data movement, state mutation, and automation risk;

\textbf{Case 3: SSL can understate source-level context.} The \texttt{electric-proxy-auth} skill is a counterexample. Its gold labels are $(\textit{risk},\textit{no},\textit{risk},\textit{no},\textit{no},\textit{risk})$. The full \texttt{SKILL.md} prediction correctly detects data-exfiltration and credential-access risk, but \texttt{SKILL.md} + SSL predicts no risk on those dimensions. The source text explicitly discusses server-side proxying, auth token injection, tenant isolation, and secret handling, while the structured record emphasizes validation and request construction. This case illustrates a current limitation: SSL helps when the normalizer faithfully surfaces resource and data-flow evidence, but it can hurt when important security semantics remain clearer in the original source prose than in the normalized fields.

\end{document}